\documentclass[letterpaper]{article} 
\usepackage{aaai2026}  
\usepackage{times}  
\usepackage{helvet}  
\usepackage{courier}  
\usepackage[hyphens]{url}  
\usepackage{graphicx} 
\usepackage{natbib}  
\usepackage{caption} 
\usepackage{algorithm}
\usepackage{algorithmic}
\usepackage{amsmath}
\usepackage{graphicx} 
\usepackage{amsfonts} 
\usepackage{booktabs}
\usepackage{newfloat}
\usepackage{listings}
\DeclareCaptionStyle{ruled}{labelfont=normalfont,labelsep=colon,strut=off} 
\floatstyle{ruled}
\newfloat{listing}{tb}{lst}{}
\floatname{listing}{Listing}
\title{MGTS-Net: Exploring Graph-Enhanced Multimodal Fusion for Augmented Time Series Forecasting}

\author{
    Shule Hao\textsuperscript{\rm 1},
    Junpeng Bao\textsuperscript{\rm 1}\thanks{Corresponding author. Email: baojp@mail.xjtu.edu.cn},
    Wenli Li\textsuperscript{\rm 2}
}

\affiliations{
    \textsuperscript{\rm 1}School of Computer Science and Technology, Xi'an Jiaotong University\\
    Xi'an, Shaanxi 710049, China\\
    haoshule@stu.xjtu.edu.cn, baojp@mail.xjtu.edu.cn\\
    \textsuperscript{\rm 2}Unmanned System Research Institute, Northwestern Polytechnical University\\
    Xi'an, Shaanxi 710072, China\\
    2024264797@mail.nwpu.edu.cn\\
}

\usepackage{bibentry}
\begin{document}
\maketitle

\begin{abstract}
Recent research in time series forecasting has explored integrating multimodal features into models to improve accuracy. However, the accuracy of such methods is constrained by three key challenges: inadequate extraction of fine-grained temporal patterns, suboptimal integration of multimodal information, and limited adaptability to dynamic multi-scale features. To address these problems, we propose MGTS-Net, a Multimodal Graph-enhanced Network for Time Series forecasting. The model consists of three core components: (1) a Multimodal Feature Extraction layer (MFE), which optimizes feature encoders according to the characteristics of temporal, visual, and textual modalities to extract temporal features of fine-grained patterns; (2) a Multimodal Feature Fusion layer (MFF), which constructs a heterogeneous graph to model intra-modal temporal dependencies and cross-modal alignment relationships and dynamically aggregates multimodal knowledge; (3) a Multi-Scale Prediction layer (MSP), which adapts to multi-scale features by dynamically weighting and fusing the outputs of short-term, medium-term, and long-term predictors. Extensive experiments demonstrate that MGTS-Net exhibits excellent performance with light weight and high efficiency. Compared with other state-of-the-art baseline models, our method achieves superior performance, validating the superiority of the proposed methodology.
\end{abstract}

\begin{links}
\end{links}

\section{1.Introduction}

Time series forecasting is pivotal in practical scenarios such as meteorological forecasting, traffic flow management, and energy consumption planning. The core challenge is twofold: accurately capturing complex patterns inherent in time series (e.g., short-term fluctuations, medium-term trends, and long-term periodicity) and integrating external factors that may alter underlying dynamics.

In practice, time series data rarely exist in isolation, often accompanied by multimodal complementary information. For instance, meteorological forecasting requires combining temperature time series with textual descriptions like "the arrival of cold waves"; traffic flow forecasting needs linking with road condition images and accident bulletin texts. Such multi-source information offers potential to improve prediction accuracy, yet effective integration remains a pressing issue.

Recent years have witnessed efforts to enhance time series forecasting via multimodal fusion. Methods like Time-LLM \cite{jin2023time} and UniTime \cite{liu2024unitime} leverage large language models (LLMs) to map time series into textual representations, bridging the language modality. Time-VLM \cite{zhong2025time}, by contrast, encodes time series into images and achieves cross-modal alignment using vision-language models. Despite advancing the field, these methods have two key limitations: (1) \textit{Modality gap}: Inherent differences between continuous time series and discrete text/image modalities cause information loss during feature mapping (e.g., mismatched time granularity between time series segments and image patches); (2) \textit{Inadequate fine-grained pattern capture}: Pre-trained models' general knowledge struggles to adapt to time series' fine-grained dynamics (e.g., minute-level fluctuations or periodic details), limiting the learning of subtle patterns.

Moreover, time series' multi-scale characteristics exacerbate forecasting difficulty. Short-term fluctuations (e.g., minute-level traffic spikes), medium-term trends (e.g., daily commuting peaks), and long-term cycles (e.g., seasonal energy consumption variations) contain complementary information. However, traditional models mostly adopt fixed-scale prediction heads and fail to dynamically adjust each scale's importance based on input sequences. For example, post-unexpected events, focus on short-term heads is needed to capture fluctuations; for stable periodic sequences, enhancing long-term heads' periodic modeling is critical. Such fixed-scale strategies significantly restrict adaptability to complex scenarios.
To address these gaps, this paper proposes MGTS-Net (Multimodal Graph-Enhanced Time-Series Network), a unified prediction framework integrating visual, textual, and time-series modalities. It dynamically models cross-modal interactions and multi-scale features via graph structures for accurate predictions. The core idea is: convert multimodal information into heterogeneous graph nodes, capture intra-modal dependencies and cross-modal correlations through relation-aware graph learning, and generate final predictions via adaptive multi-scale fusion. Specifically, MGTS-Net comprises three key components:
\begin{enumerate}
    \item \textbf{Multimodal Feature Extraction Layer}: Customized encoders are designed for time-series, image, and text modalities: For time-series encoding, a Frequency-Time Cell (FTC) enhances Time-MOE's capability to capture joint frequency-time domain features; For image encoding, asymmetric patch partitioning and temporal positional embedding optimize ViT's temporal feature extraction; For text encoding, pre-trained language models extract semantic representations of three text types (trend descriptions, variable explanations, event records), ensuring all modalities retain temporal attributes.
    \item \textbf{Multimodal Feature Fusion Layer (MFF)}: A heterogeneous graph models multiple relationships: cross-modal alignment (e.g., one-to-one mapping between time-series segments and image patches, one-to-many influence of text on all nodes); intra-modal temporal dependencies (e.g., past-future adjacency in time-series and image nodes). Graph Neural Networks aggregate neighbor information to dynamically refine time-series node features, enabling accurate multimodal knowledge enhancement.
    \item \textbf{Multi-Scale Prediction Layer (MSP)}: Three prediction heads (short-term, medium-term, long-term) capture patterns at different scales. Fusion weights are dynamically generated via MLP based on input time-series features, adaptively integrating multi-scale results to improve adaptability to complex patterns.
\end{enumerate}

Our key contributions can be summarized as follows:
\begin{itemize}
    \item Proposing the MFF module, which models cross-modal alignment and intra-modal temporal dependencies via a heterogeneous graph, enabling dynamic enhancement of time-series features with multimodal information.
    \item Improving pre-trained models' temporal adaptability: enhancing time-series encoders' joint frequency-time modeling via FTC; optimizing ViT's temporal image encoding with asymmetric patch partitioning to strengthen temporal correlation of multimodal features.
    \item Designing an adaptive multi-scale prediction mechanism that fuses multi-scale results via dynamic weights (generated by MLP from input features), effectively leveraging multi-scale patterns to boost prediction accuracy.
\end{itemize}

\section{2. Related Work}
\subsection{2.1 Unimodal Time Series Forecasting}
Unimodal methods extract patterns from pure time series, divided into statistical and deep learning models. Statistical models like ARIMA \cite{box1970time} capture linear temporal dependencies via differencing and autoregressive components but struggle with nonlinear fluctuations and long-term patterns.
Deep learning models such as LSTM \cite{NIPS1996_a4d2f0d2} and GRU \cite{cho2014learning} use gating mechanisms to model short-term dependencies. Transformer-based models \cite{vaswani2017attention} advance long-horizon forecasting via self-attention, with variants like Informer \cite{zhou2021informer}, Autoformer \cite{wu2021autoformer}, ETSFormer \cite{woo2022etsformer}, PatchTST \cite{huang2024long}, Non-Stationary Transformer \cite{liu2022non}, and TimesNet \cite{wu2022timesnet} enhancing specific capabilities.
MoE-based models like TIME-MOE \cite{shi2024time} integrate MoE with Transformers but lack frequency-domain optimization. Foundation models like Time-Mixer \cite{wang2024timemixer} and TimeGPT \cite{nixtla2023timegpt} show potential but limitations. These methods, relying solely on time series, lack external information, limiting accuracy in complex scenarios.
\subsection{2.2 Multimodal Time Series Forecasting}
Multimodal fusion incorporates text and images. Text-enhanced methods use PLMs: Time-LLM \cite{jin2023time}, GPT4MTS \cite{jia2024gpt4mts}, TimeCMA \cite{liu2024timecma}, UniTime \cite{liu2024unitime}, and TimeRAG \cite{yang2025timerag} address alignment to varying degrees.
Vision-enhanced methods include CNN-based \cite{semenoglou2023image}, Time-VLM \cite{zhong2025time}, Bi-Mamba4TS \cite{liang2024bi}, Timer-XL \cite{liu2024timer}, Lag-Llama \cite{rasul2023lag}, and TIME-MOE \cite{shi2024time}. Key bottlenecks: inconsistent temporal attributes and simplistic cross-modal modeling.
\subsection{2.3 Graph-Enhanced Time Series Models}
GNNs model relational dependencies. ST-GCN \cite{yan2018spatial}, ASTGCN \cite{zhu2021ast}, DCRNN \cite{li2017diffusion}, CGMF \cite{wang2023learning}, and FC-STGNN \cite{wang2024fully} handle dynamic dependencies but focus on single modalities.
Multimodal graph fusion (MST-GAT \cite{ding2023mst}, Spatio-Temporal Meta-Graph Learning \cite{jiang2023spatio}, \cite{zhang2022multi}) lacks visual integration or uses simple relations, failing to capture complex alignments.
\begin{figure*}[t]
\centering
\includegraphics[width=\textwidth]{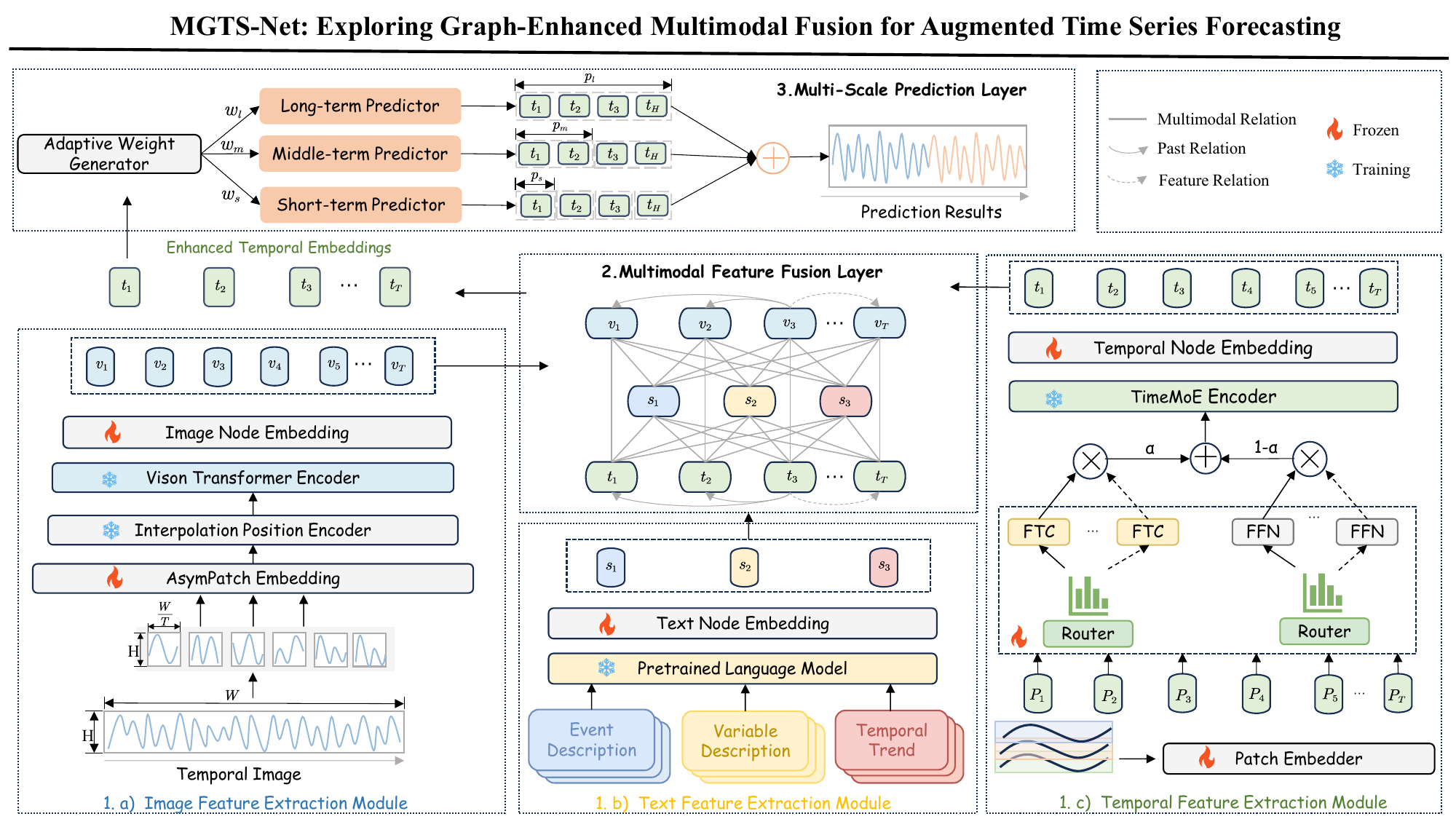} 
\caption{Overview of the MGTS-Net framework.}
\label{fig1}
\end{figure*}

\section{3. Methodology}
To address the limitations of unimodal methods and leverage the complementary advantages of visual, textual, and temporal modalities, we propose MGTS-Net, a unified framework that dynamically enhances temporal features through graph neural networks with image features and text features. As illustrated in Figure \ref{fig1}, the framework comprises three core components:
\begin{itemize}
    \item \textbf{Multimodal Feature Extraction Layer (MFE)}: It converts temporal, textual, and image data into modality-specific nodes.
    \item \textbf{Multimodal Feature Fusion layer (MFF)}: It optimizes temporal nodes through cross-modal graph interaction.
    \item \textbf{Multi-Scale Prediction layer (MSP)}: It makes predictions at multiple scales simultaneously and fuses the multi-scale prediction results through dynamic weights.
\end{itemize}

\subsection{3.1 Multimodal Feature Extraction Layer}
\subsubsection{Text Feature Extraction Module}
We define three categories of texts: time-series trend texts (e.g., describing trend features such as "first rising then falling"), variable description texts (explaining the meanings of variables in time-series data, such as definitions of wind speed and humidity), and event description texts (referring to sudden events affecting forecasting, such as regional earthquakes). Specifically, let \( T_t = \{t_1, t_2, t_3, t_4\} \) denote the set of time-series trend texts, containing four typical trends; let \( T_v = \{v_1, v_2, \dots, v_V\} \) represent variable description texts, where the \( v \)-th text corresponds to the \( v \)-th variable in the time series; and let \( T_e = \{e_1, e_2, \dots, e_E\} \) denote event descriptions (possibly empty), derived from multi-source real-time information. These are integrated into a text collection \( P_j = \{w_{j,1}, w_{j,2}, \dots, w_{j,l_j} \mid j = 1, 2, \dots, N\} \), where \( l_j \) is the length of the \( j \)-th text.
We use pre-trained language models to map each text \( T \) into a \( d \)-dimensional embedding. Taking the widely adopted encoder model BERT as an example:
\begin{equation}
\scalebox{0.8}{\(    e_j = \textit{BERT}([\text{CLS}], w_{j,1}, w_{j,2}, \dots, w_{j,l_j}, [\text{SEP}]) \in \mathbb{R}^{(l_j + 2) \times d}
\)}
\end{equation}
where \([\text{CLS}]\) and \([\text{SEP}]\) are special tokens marking the sequence start and end. BERT generates embeddings for each input token, and the \([\text{CLS}]\) position captures the overall semantic information. We thus take its hidden state as the text's semantic representation: \( E_j = e_j([\text{CLS}]) \), forming the semantic matrix \( E_{\text{text}} = [E_1, E_2, \dots, E_N] \in \mathbb{R}^{N \times d} \).
into overlapping patches with length \( pl \) and stride \( st \),

\subsubsection{Temporal Feature Extraction Module}
Let \( x_{\text{enc}} \in \mathbb{R}^{B \times L \times D} \) denote the input time series, where \( B \), \( L \), and \( D \) are the batch size, sequence length, and number of variables, respectively. We first split \(x_{\text{enc}}\) into overlapping patches with length \( pl \) and stride \( st \). . Each patch is linearly projected into a \( d \)-dimensional latent space, with positional embeddings added to preserve temporal order, resulting in patch embeddings \( E_{\text{temporal}} \in \mathbb{R}^{B \times T \times d} \), where the total number of patches is \( T = (L - pl)/st + 1 \). These temporal patch embeddings are then fed into an improved TIME-MOE module, with modifications fine-tuned accordingly.

In the original MoE Transformer Block of TIME-MOE \cite{shi2024time}, the Mixture-of-Experts (MoE) layer lacks explicit modeling of frequency-domain periodic features. To enhance the model's capability in capturing joint time-frequency domain features, we integrate a Frequency-Time Cell (FTC)\cite{liu2025mofe} into selected MoE Transformer Blocks, enabling simultaneous modeling of both frequency-domain and time-domain characteristics. The core process is as follows:
\begin{align}
    u_t^l =& \text{SA}(\text{RMSNorm}(h_t^{(l-1)})) + h_t^{(l-1)}\\
    \bar{u}_t^l =& \text{RMSNorm}(u_t^l)\\
    h_t^l =& \text{FTCMoE}(\bar{u}_t^l) + u_t^l
\end{align}

where \( \text{SA} \) denotes self-attention with a causal mask, and \( \text{FTCMoE} \) refers to the MoE layer integrated with FTC. The expert module is defined as:

\begin{equation}
    \text{Expert}_i(x) = \alpha \cdot \text{FTC}(x) + (1 - \alpha) \cdot \text{FFN}(x)
\end{equation}

where \( \alpha \in [0, 1] \) is a learnable fusion coefficient, dynamically balancing the weights of frequency-domain and time-domain features.

Specifically, the FTC operates as follows: the input\(x\) is split into a frequency-domain branch \( x_f \) and a time-domain branch \( x_t \). The frequency-domain branch is mapped to the frequency domain via linear transformation and converted back to the time domain using harmonic basis functions:

\begin{equation}
    x_f = \text{Linear}(x; W_f), \quad x_{t,\text{freq}} = e^{j \cdot x_f}
\end{equation}

where \(j\) is the imaginary unit, and \( e^{j \cdot x_f} \) is a Fourier-based harmonic function. The time-domain signal \( x_{t,\text{freq}} \) from the frequency-domain transformation is concatenated with the original time-domain branch \( x_t \), i.e., \( \text{FTC}(x) = \text{Linear}([x_t, x_{t,\text{freq}}]) \).

We retain the original Top-K routing strategy of MoE, but the routing weights also consider the importance of frequency-domain features:

\begin{equation}
    g_{i,t} = 
\begin{cases} 
s_{i,t}, & \text{if } s_{i,t} \in \text{TopK}(\{s_{j,t}\}, K), \\
0, & \text{otherwise},
\end{cases}
\end{equation}

where \( s_{i,t} \) is the score of expert \(i\) for the input \( x_t \), computed by a router that includes frequency-domain features:

\begin{equation}
    s_{i,t} = \text{Softmax}(\text{Linear}(\text{FTC}(x_t) + x_t))
\end{equation}

Thus, the proposed TimeFTCMoE model captures richer temporal features.

\subsubsection{Image Feature Extraction Module}
We adopt an asymmetric partitioning mode to reshape the image into a flattened sequence of 2D temporal image patches: \( x_p \in \mathbb{R}^{T \times P \times C} \), where each image patch has a resolution of \( P = (H, W/T) \), and \( T \) serves as the effective input sequence length of the Transformer, matching the number of temporal patches. Each image patch is mapped to a feature vector via convolution: \( x_i = \text{Conv2d}(x_p) \in \mathbb{R}^D \).

To adapt to this partitioning mode, we use bicubic interpolation to adjust the original 14×14 grid positional encoding to a \( (1, T)\) To adapt to this partitioning mode, we use bicubic interpolation to adjust the original 14×14 grid positional encoding to a \( T \) temporal image patches:
\begin{equation}
    \text{NPE} = \text{interpolate}(\text{PE}, \text{size} = (1, N))
\end{equation}

In the temporal positional encoding fusion stage, we introduce positional encoding adapted to the \( T \) image patches, yielding the fused input sequence: \( z_i = x_i + \text{NPE}_i \). The sequence \( \{z_1, z_2, \dots, z_N\} \) is fed into the Transformer encoder, which processes it through stacked multi-layer self-attention mechanisms and feed-forward networks to output an image representation vector containing global temporal features:
\begin{equation}
    \text{E}_{\text{image}} = \text{Transformer}(\{z_1, z_2, \dots, z_N\}) \in \mathbb{R}^{T \times d}
\end{equation}
\begin{algorithm}[tb]
\caption{Multi-Scale Prediction Algorithm }
\label{alg:multiscale}
\textbf{Input}: Target output length $H$ (e.g., 96), hidden state $h_t^L \in \mathbb{R}^D$ from last Transformer block\\
\textbf{Parameter}: Forecast horizons of predictors $\{s=30, m=50, l=100\}$ (short, medium, long-term)\\
\textbf{Output}: List of truncated predictions $\text{Results} = [\hat{X}_s^H, \hat{X}_m^H, \hat{X}_l^H]$
\begin{algorithmic}[1] 
\STATE Initialize $\text{Results} = []$ and $\text{scales} = [s, m, l]$.
\FOR{each scale in $\text{scales}$}
    \STATE Calculate iterations: $\text{iters} = \lceil H / \text{scale} \rceil$.
    \STATE Generate full prediction via corresponding predictor: 
    \STATE \quad $\text{full\_pred} = \text{Predictor}(h_t^L, \text{iters})$.
    \STATE Truncate to target length: $\text{truncated\_pred} = \text{full\_pred}[1:H]$.
    \STATE Append to results: $\text{Results.append}(\text{truncated\_pred})$.
\ENDFOR
\STATE \textbf{return} $\text{Results}$
\end{algorithmic}
\end{algorithm}
\subsection{3.2 Multimodal Feature Fusion layer}
The node set \( V \) includes three types corresponding to text, time-series, and image modalities. Relation types \( R \) are categorized into three classes to model cross-modal associations and temporal dependencies:

\subsubsection{Multimodal Relations} Capture semantic correlations between different modalities. For time-series and image nodes, as they are segmented into one-to-one correspondences across the same temporal length, each time-series node \( u_i^t \) connects to its corresponding image node \( u_i^i \). The edge set is \( E_{u^t-u^i} = \{(u_i^t, u_i^i) \mid i = 1, 2, \dots, n\} \) (where \( T = \{u_1^t, u_2^t, \dots, u_i^t\} \) is the time-series node set and \( I = \{u_1^i, u_2^i, \dots, u_i^i\} \) is the image node set). For text nodes and other modalities, text nodes (describing trends, variables, or events) provide context for arbitrary time intervals and their visualizations. Thus, each text node connects to all time-series and image nodes, with edge sets \( E_{s-t} = \{(u_j^s, u_i^t) \mid j = 1, 2, \dots, m; i = 1, 2, \dots, n\} \) (where \( S = \{u_1^s, u_2^s, \dots, u_i^s\} \) is the text node set) and \( E_{s-i} = \{(u_j^s, u_i^i) \mid j = 1, 2, \dots, m; i = 1, 2, \dots, n\} \).

\subsubsection{Past Relations} Model temporal dependencies within the same modality by connecting nodes to their preceding neighbors. For time-series nodes, node \( u_i^t \) connected to preceding nodes \( u_j^t \) (\( j < i \)) within the time window \( w_p \), forming the edge set:
\begin{equation}
     E_{u_p^t} = \{(u_j^t, u_i^t) \mid j \in [i - w_p, i - 1], i = 1, 2, \dots, n\}
\end{equation}
  
   For image nodes, node \( u_i^i \) connected to preceding image nodes \( u_j^i \) (\( j < i \)) in the corresponding temporal order, forming the edge set:
\begin{equation}
    E_{u_p^i} = \{(u_j^i, u_i^i) \mid j \in [i - w_p, i - 1], i = 1, 2, \dots, n\} 
\end{equation}

   Text nodes lack past relations as their descriptions do not follow a strict temporal order.

\subsubsection{Future Relations} Complement past relations by connecting nodes to their subsequent neighbors within the same modality. For time-series nodes, node \( u_i^t \) connected to subsequent nodes \( u_j^t \) (\( j > i \)) within the future window \( w_f \), forming the edge set:
\begin{equation}
\scalebox{0.85}{$
    E_{u_f^t} = \{(u_i^t, u_j^t) \mid j \in [i + 1, i + w_f], i = 1, 2, \dots, n - w_f\}
$}
\end{equation}
   For image nodes, node \( u_i^i \) is connected to subsequent image nodes \( u_j^i \) (\( j > i \)), forming the edge set:
\begin{equation}
\scalebox{0.85}{$
    E_{u_f^i} = \{(u_i^i, u_j^i) \mid j \in [i + 1, i + w_f], i = 1, 2, \dots, n - w_f\}
$}
\end{equation}
   Text nodes similarly lack future relations for the same reason as past relations.

For each relation type \( r \in R \), node features are updated by aggregating neighbor information via relation-specific weight 
\begin{table*}[!ht]
    \centering
    \begin{tabular}{c|c|c|c|c|c|c|c|c|c|c}
        \toprule
        Methods & MGTS-Net  & Time-VLM  & MoFE-time & Time-LLM & TimeMoE & TimesNet & PatchTST & DLinear    \\ 
        \midrule
        Metric  & MSE  MAE & MSE MAE & MSE MAE & MSE MAE & MSE  MAE & MSE MAE & MSE MAE & MSE MAE  \\ 
        \midrule
        {\textit{ETTh1}} & \textbf{0.389 0.412} & 0.405 0.420 & 0.396 0.422 & 0.431 0.456 & 0.403 0.430 & 0.458 0.450 & 0.413 0.430 & 0.442 0.450  \\ 
        \midrule
        {\textit{ETTh2}}& \textbf{0.308 0.366} & 0.341 0.391 & 0.438 0.439 & 0.353 0.396 & 0.472 0.458 & 0.414 0.427 & 0.330 0.379 & 0.363 0.417  \\ 
        \midrule
        {\textit{ETTm1}}& \textbf{0.312 0.371} & 0.350 0.377 & 0.391 0.420 & 0.356 0.396 & 0.407 0.427 & 0.400 0.406 & 0.351 0.380 & 0.653 0.550  \\ 
        \midrule
        {\textit{ETTm2}}& \textbf{0.238 0.298} & 0.248 0.311 & 0.278 0.347 & 0.291 0.341 & 0.324 0.377 & 0.291 0.333 & 0.255 0.315 & 0.292 0.376  \\ 
        \midrule
        {\textit{Weather}}& \textbf{0.217 0.260} & 0.224 0.263 & 0.229 0.271 & 0.231 0.278 & 0.245 0.285 & 0.259 0.287 & 0.225 0.264 & 0.240 0.304  \\ 
        \midrule
        {\textit{Electricity}} & \textbf{0.156 0.251} & 0.172 0.273 & 0.203 0.301 & 0.186 0.294 & 0.212 0.311 & 0.192 0.295 & 0.161 0.252 & 0.223 0.320 \\
        \midrule
        {\textit{Traffic}} & \textbf{0.381 0.261} & 0.419 0.303 & 0.502 0.370 & 0.469 0.327 & 0.517 0.384 & 0.620 0.336 & 0.390 0.263 & 0.638 0.452 \\
        \midrule
        {\textit{Avg}} & \textbf{0.286 0.317} & 0.308 0.334 & 0.348 0.367 & 0.331 0.358 & 0.369 0.382 & 0.392 0.362 & 0.304 0.326 & 0.407 0.410\\
        \bottomrule
    \end{tabular}
    \caption{Results are averaged over forecasting horizons \( H \in \{96, 192, 336, 720\} \). A lower value indicates better performance. Full results see appendix.}
    \label{tab1}
\end{table*}

matrices to capture local dependencies (e.g., cross-modal interactions between time-series and image nodes, short-term trends among adjacent time-series nodes). The feature update formula for node \( v_i \) at layer \( k \) is:
\begin{equation}
\scalebox{0.85}{$
    h_i^{(k)} = \sigma\left( \sum_{r \in R} \sum_{j \in N_r(i)} \frac{1}{c_{ir}} W_r^{(k)} h_j^{(k-1)} + W_0^{(k)} h_i^{(k-1)} \right)
$}
\end{equation}
where \( \sigma \) is the ReLU activation function, \( N_r(i) \) is the set of neighbors connected to node \( i \) via relation \( r \), \( c_{ir} \) is a normalization constant ensuring consistent aggregation across relation types, and \( W_r^{(k)} \) and \( W_0^{(k)} \) are the weight matrices for relation \( r \) and self-connections at layer \( k \), respectively. The resulting representation of the \( i \)-th utterance under modality \( m \in \{t, i, s\} \) is denoted as \( h_i^m \). Time-series nodes are extracted and rearranged chronologically to form a continuous temporal feature sequence \( h_t \in \mathbb{R}^{B \times T \times d} \), which is fed into the multi-scale prediction module for subsequent forecasting.

\subsection{3.3 Multi-Scale Prediction layer}

In time series forecasting, patterns at different temporal scales (e.g., short-term fluctuations, medium-term trends, and long-term periodicity) often contain complementary information. To fully leverage these multi-scale features, we propose a multi-scale predictor that dynamically integrates predictions from different scales through an adaptive weight fusion mechanism. Given the hidden representation of the input sequence 
input sequence \( h_t \in \mathbb{R}^{B \times T \times d} \), three dedicated prediction heads are designed to address distinct temporal scales: the short-term head focuses on capturing high-frequency patterns, the medium-term head targets trend changes over intermediate cycles, and the long-term head models persistent trends and periodicity. The prediction horizons for these heads are set as \( p = \{p_s, p_m, p_l\} \) time steps. The prediction process is formalized as follows:

\begin{equation}
    \hat{X}_p^{(i)} = W_p \cdot h_{t + p(i-1)} \quad \text{for} \quad i = 1, 2, \dots, n
\end{equation}

where \( W_p \in \mathbb{R}^{p \times D} \) is a learnable parameter matrix, and \( h_{t + p(i-1)} \) denotes the hidden state at the \( i \)-th iteration. After \( n \) iterations, \( n \times P \) time steps of predictions are generated, which are then truncated to the target horizon \( H \), resulting in a prediction sequence denoted as \( \hat{X}_p^H \). As shown in Algorithm~\ref{alg:multiscale}.

To dynamically balance the contributions of these scale-specific heads, adaptive weights are computed based on the temporal features:

\begin{equation}
    W = \text{Softmax}(\text{MLP}(h_t))
\end{equation}

where the MLP consists of two linear transformations with ReLU activations, and outputs a weight distribution \( w \in \mathbb{R}^{B \times M} \) corresponding to \( M \) prediction heads. The final prediction is obtained by weighted fusion of individual head outputs:

\begin{equation}
    \hat{Y} = \sum_{m=1}^{M} w_m \cdot \hat{X}_H^P
\end{equation}

This mechanism enables the model to dynamically adjust the importance of each scale based on input characteristics (e.g., volatility, trend strength), thereby maximizing the complementary value of multi-scale information.

\begin{table*}[h]
    \centering
    \begin{tabular}{c|c|c|c|c|c|c|c|c|c|c}
        \toprule
        Methods & MGTS-Net  & Time-VLM  & MoFE-time & Time-LLM & TimeMoE & TimesNet & PatchTST & DLinear    \\ 
        \midrule
        Metric  & MSE  MAE & MSE MAE & MSE MAE & MSE MAE & MSE  MAE & MSE MAE & MSE MAE & MSE MAE  \\ 
        \midrule
        {\textit{ETTh1}} & \textbf{0.402 0.422} & 0.442 0.453 & 0.447 0.480 &	0.627 0.543 & 0.445 0.472 & 0.925 0.647 & 0.694 0.569 & 0.750 0.611 \\ 
        \midrule
        {\textit{ETTh2}}& \textbf{0.326 0.397} & 0.354 0.402 & 0.450 0.455 & 0.382 0.418 & 0.410 0.444 & 0.439 0.448 & 0.827 0.615 & 0.694 0.577  \\ 
        \midrule
        {\textit{ETTm1}}& \textbf{0.333} 0.388 &	0.364 \textbf{0.385} & 0.433 0.4449 & 0.425 0.434 & 0.501 0.418 & 0.717 0.561 & 0.526 0.476 & 0.400 0.417  \\ 
        \midrule
        {\textit{ETTm2}}& \textbf{0.248 0.309} & 0.262 0.323 & 0.279 0.347 & 0.299 0.343 & 0.334 0.379 & 0.344 0.372 & 0.314 0.352 & 0.399 0.426  \\ 
        \midrule
        {\textit{Weather}}& \textbf{0.222 0.270} & 0.240 0.280 &	0.248 0.284 & 0.260 0.309 & 0.255 0.295 & 0.298 0.318 & 0.269 0.303 & 0.263 0.308  \\ 
        \midrule
        {\textit{Electricity}} & \textbf{0.176 0.301} & 0.218 0.315 & 0.213 0.315 & 0.195 0.300 & 0.190 0.285 & 0.402 0.453 & 0.181 0.277 & 0.276 0.375 \\
        \midrule
        {\textit{Traffic}} & \textbf{0.403 0.289} & 0.558 0.410 & 0.550 0.407 & 0.560 0.415 & 0.601 0.396 & 0.867 0.493 & 0.418 0.296 & 0.450 0.317 \\
        \midrule
        {\textit{Avg}} & \textbf{0.301 0.342} & 0.348 0.367 & 0.360 0.389 & 0.392 0.397 & 0.391 0.384 & 0.570 0.467 & 0.461 0.413 & 0.447 0.419\\
        \bottomrule
    \end{tabular}
    \caption{Few-shot learning on 5\% training data. Results are averaged over forecasting horizons \( H \in \{96, 192, 336, 720\} \). Lower values indicate better performance. Full results see appendix.}
    \label{tab2}
\end{table*}
\begin{table}[h]
    \centering
    \scriptsize  
    \setlength{\tabcolsep}{3pt}  
    \begin{tabular}{c|c|c|c|c|c}
        \toprule
        Methods & MGTS-Net & w/o MGET & w/o MSP & w/o MGTS-VIT & w/o FTC \\ 
        \midrule
        Metric  & MSE  MAE & MSE MAE & MSE MAE & MSE MAE & MSE  MAE \\ 
        \midrule
        {\textit{ETTh1}} &\textbf{0.389 0.412} & 0.430 0.441 & 0.401 0.422 & 0.423 0.436 & 0.395 0.418 \\ 
        \midrule
        {\textit{Weather}}& \textbf{0.217 0.260} & 0.232 0.275 & 0.225 0.265 & 0.228 0.274 & 0.218 0.263 \\ 
        \midrule
        {\textit{Electricity}} & \textbf{0.156 0.251} & 0.209 0.301 & 0.177 0.273 & 0.201 0.298 & 0.170 0.264 \\
        \midrule
        {\textit{Traffic}} & \textbf{0.381 0.261} & 0.409 0.301 & 0.394 0.283 & 0.405 0.297 & 0.393 0.280 \\
        \midrule
        {\textit{Avg}} & \textbf{0.286 0.296} & 0.320 0.330 & 0.300 0.311 & 0.314 0.326 & 0.293 0.306 \\
        \bottomrule
    \end{tabular}
    \caption{Ablation studies on ETTh1, Weather, Electricity and Traffic by removing/replacing key modules.}
    \label{tab3}
\end{table}

\section{4. Experiments}

\subsubsection{Datasets metrics} The datasets used in our experiments are widely recognized time-series benchmark datasets, such as those detailed in \cite{zhou2021informer}. These datasets include weather, traffic, electricity, and four ETT datasets. Performance is measured using MAE and MSE, following the standard evaluation practices in this field. Detailed statistics about these datasets are provided in appendix.

\subsubsection{Baselines} To comprehensively demonstrate the performance of our model, we will compare it with state-of-the-art time series models. If applicable, we refer to their performance reported in \cite{zhong2025time}. Our baseline models include multimodal methods (e.g. Time-VLM \cite{zhong2025time}, Time-LLM \cite{jin2023time}, GPT4TS \cite{zhou2023one}, and TimesNet \cite{wu2022timesnet}); traditional deep models (e.g., PatchTST \cite{huang2024long}, DLinear \cite{zeng2023transformers}); and MOE models (e.g., MoFE-time \cite{liu2025mofe} and Time-MoE \cite{shi2024time}).
\subsubsection{Implementation Details} We use a unified evaluation pipeline and compare MGTS-Net with excellent baselines according to the configuration of \cite{wu2022timesnet} to ensure fairness. Among them, Bert \cite{devlin2019bert} and Vision Transformer \cite{dosovitskiy2020image} are selected as the language model and visual model respectively. All models are trained using the Adam optimizer (initial learning rate, halved per epoch), with a batch size of 32 for a maximum of 10 epochs, and an early stopping mechanism is employed. The experiments are run on an Nvidia RTX A6000 GPU (48GB). See the appendix for more details.

\subsection{4.1 Long-term Forcasting}

\subsubsection{Setting.} We evaluate the long-term forecasting capabilities
of MGTS-Net across multiple horizons and datasets.
\begin{figure}[h]
    \centering
    \includegraphics[width=0.45\textwidth]{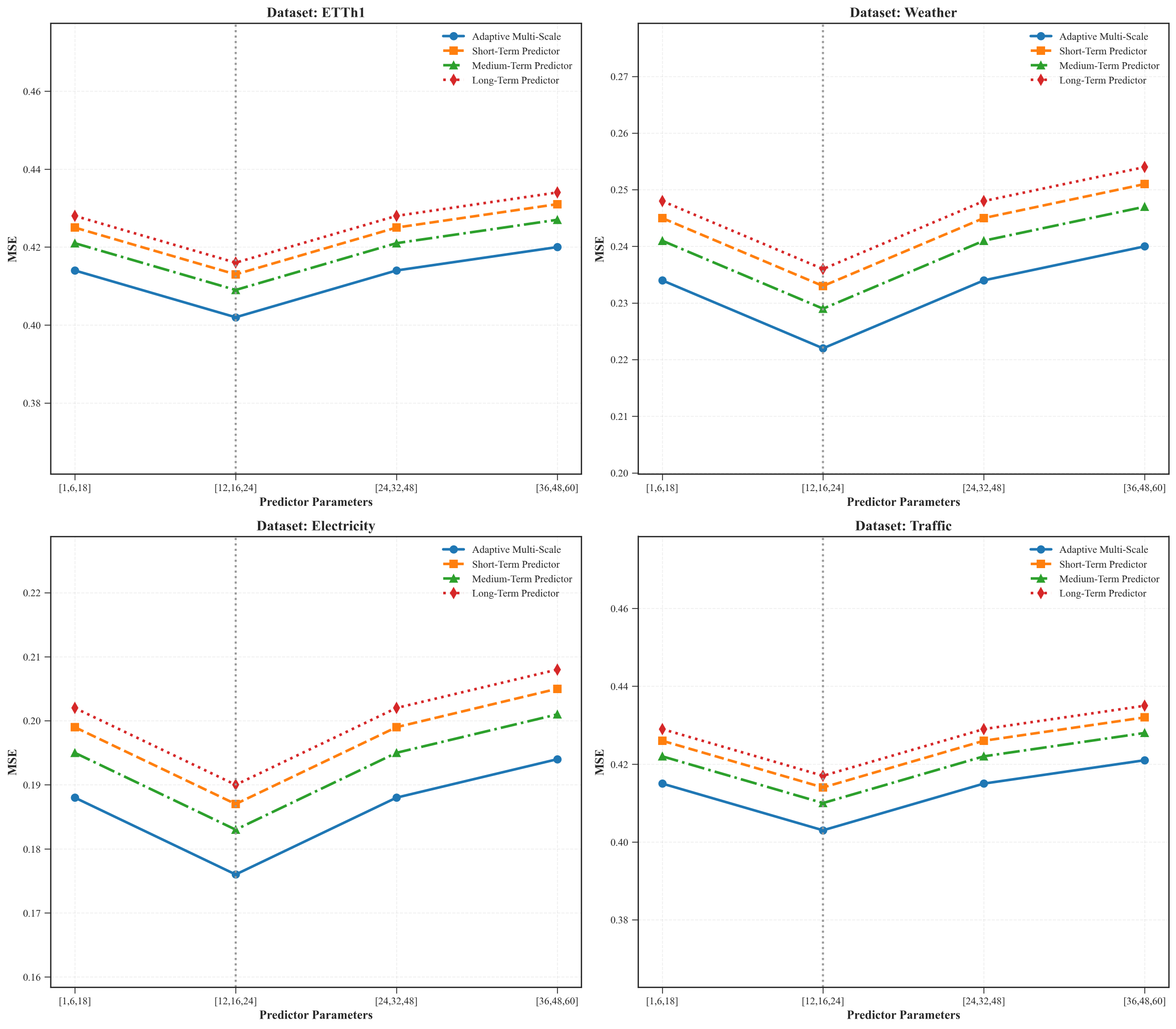}
    \caption{Results of averaging the parameter results of different prediction heads over the forcasting horizons \( H \in \{96, 192, 336, 720\} \)}
    \label{fig2}
\end{figure}
\subsubsection{Results.} As shown in Table \ref{tab1}. MGTS-Net outperforms all baselines in most scenarios, with significant advantages over key competitors. Compared to Time-VLM (a sota multimodal model), MGTS-Net achieves an average improvement of 1.95\% and 5.6\%. Against Time-LLM (text-enhanced, using BERT as the backbone), MGTS-Net reduces MSE by an average of 0.045. It also outperforms traditional models like DLinear by a large margin, confirming the effectiveness of its graph-enhanced multimodal fusion.

\subsection{4.2 Few-shot Forcasting}
\subsubsection{Setting.} To evaluate the performance of MGTS-Net under few-shot conditions, the same setup as Time-VLM was adopted, and only 5\% of the training time steps were used for model training.

\subsubsection{Results.} Table \ref{tab2} shows that MGTS-Net and Time-VLM (both fusing text, image, and time series modalities) outperform unimodal baselines, verifying multimodal fusion’s value in sparse data settings. MGTS-Net consistently outperforms Time-VLM, highlighting the superiority of its graph-based fusion in capturing cross-modal associations and temporal dynamics, thus validating the effectiveness of its multimodal strategy.
\begin{figure}[h]
    \centering
    \includegraphics[width=0.45\textwidth]{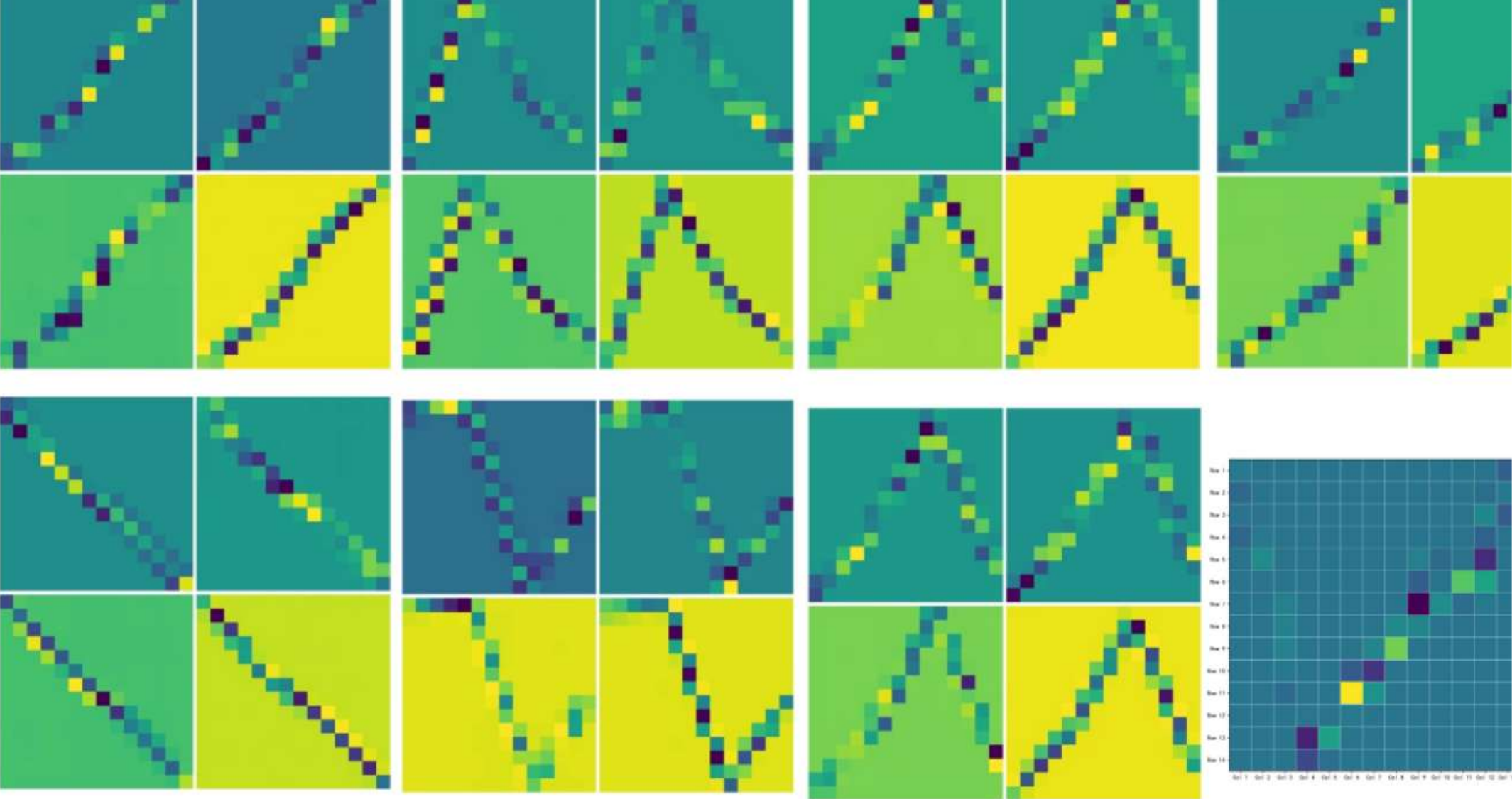}
    \caption{Visual comparison of image feature extraction between MGTS-Net and VIT: The first seven blocks represent the features of each image patch, and the last block shows the features of 14x14 patches in the VIT model.}
    \label{fig3}
\end{figure}

\subsection{4.3 Model Analysis}
\subsubsection{Ablation Studies.} To validate component contributions, we conduct ablation studies on ETTh1, Weather, and Electricity by removing/replacing key modules: \textbf{w/o MFF}: Replace the multimodal feature fusion layer with simple concatenation; 
\textbf{w/o MSP}: Use a single MLP head instead of adaptive multi-scale prediction; 
\textbf{w/o MGTS-ViT}: Replace customized MGTS-ViT with the original ViT; 
\textbf{w/o FTC}: Remove the frequency-time cell (FTC) experts from the Time-MoE encoder and only retain the time-domain feature modeling ability.
We also test four sets of prediction horizon parameters to analyze scale impacts.

The results in Table \ref{tab3} show that the full MGTS-Net outperforms all ablated versions, confirming the necessity of each module. The most significant performance drop occurs in \textbf{w/o MFF}, indicating traditional concatenation fails to capture deep cross-modal dependencies, validating the irreplaceable role of graph-structured MFF. Removing MGTS-ViT also degrades performance, highlighting the limitations of original ViT for sequential line charts and the value of custom patch partitioning.
As shown in Figure \ref{fig2}, the adaptive multi-scale prediction layer (MSP) outperforms fixed-scale alternatives across parameters, further verifying its effectiveness in dynamic fusion. These results confirm the synergistic contributions of all components, validating the rationality of the architecture.

\subsubsection{Computation Studies.} MGTS-Net exhibits strong computational efficiency (Table \ref{tab4hsl}), with 37.3M trainable parameters and low memory usage. Text data is stored on disk and loaded on-demand, enabling flexible adaptation to diverse datasets. Compared to MoE-based models like Time-MoE, MGTS-Net achieves faster inference, efficiently handling dynamic loads—highlighting its lightweight design and scalability for real-world deployment.

\subsubsection{Visualization.} To analyze multimodal temporal modeling, we visualize the processes of feature capture and cross-modal fusion as follows:

Temporal image features: As shown in Figure \ref{fig3}: Native ViT shows scattered responses, while MGTS-Net’s graph-enhanced module reveals clear trends (e.g. periodic peaks/valleys) in heatmaps, confirming improved dynamic pattern capture.

\begin{figure}[h]
    \centering
    \includegraphics[width=0.45\textwidth]{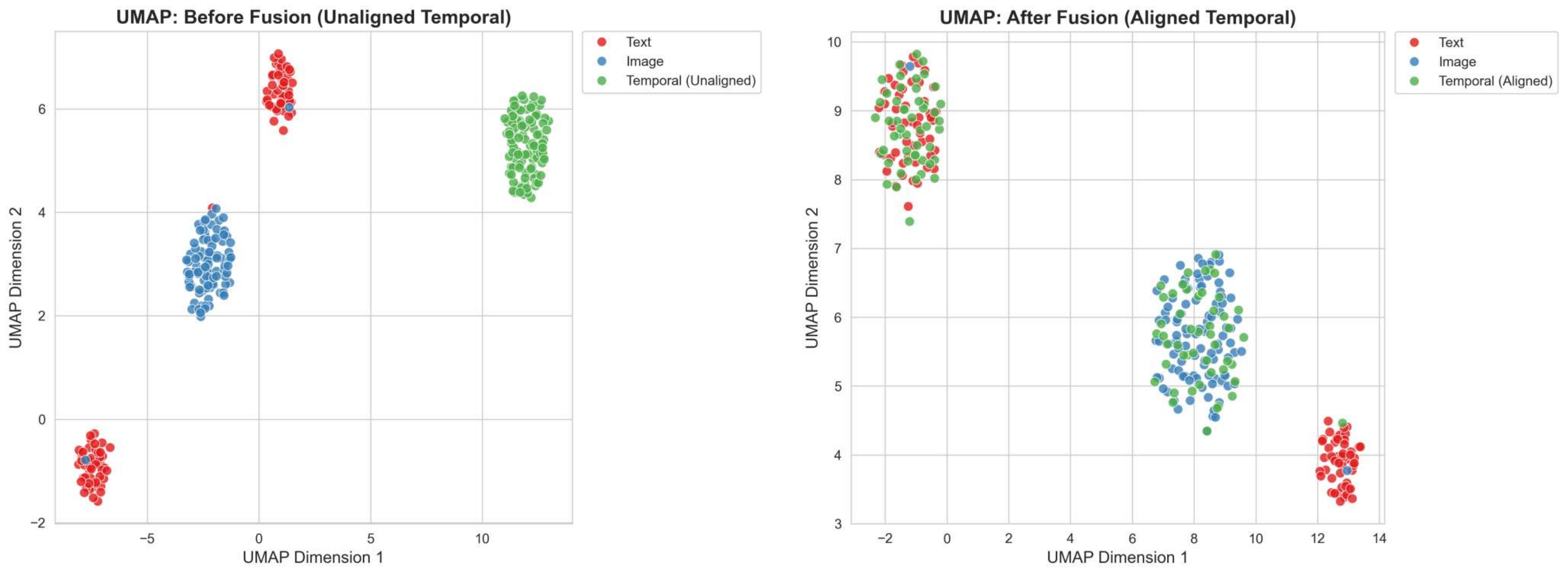}
    \caption{Interpretability visualization of MGTS-Net: multimodal feature alignment via UMAP.}
    \label{fig4}
\end{figure}

UMAP visualization: Before fusion, text, image, and temporal features form isolated clusters. After fusion, temporal features align closely with text/image clusters (Figure \ref{fig4}), demonstrating effective cross-modal learning and feature space unification—validating MFF’s ability to align multimodal information.
\begin{table}[h]
    \centering
    \scriptsize  
    \setlength{\tabcolsep}{3pt}  
    \begin{tabular}{c|c|c|c|c|c}
        \toprule
        Methods & MGTS-Net & Time-VLM & Time-LLM & Time-MoE & MoFE-Time \\ 
        \midrule
        Train Param.(M) & \textbf{37.3} & 143.6 & 51.0509	& 113.35 & 117.95 \\ 
        \midrule
        Mem.(MiB) &	\textbf{2060} & 2630 & 27723 & 2090 & 2070 \\ 
        \midrule
        Speed(s/iter) & \textbf{0.115} &	0.481 &	1.327 &	0.120 &	0.130 \\ 
        \bottomrule
    \end{tabular}
    \caption{Efficiency analysis when forecasting on the ETTh1 dataset.}
    \label{tab4hsl}
\end{table}

\section{5.Conclusion}

This paper proposes MGTS-Net, a unified multimodal framework for time series forecasting. By optimizing modality-specific encoders, incorporating frequency-time cells for enhanced joint time-frequency modeling and refining ViT via temporal-aware partitioning, MGTS-Net captures fine-grained temporal features across time series, images, and text.
A heterogeneous graph models intra-modal dependencies and cross-modal alignment, enabling dynamic multimodal knowledge aggregation. Adaptive fusion of multi-scale prediction layer extends multi-resolution design principles, balancing short-term fluctuations and long-term periodicity. MGTS-Net operates on raw time series alone, achieving self-enhancement via internal visual/textual generation—addressing auxiliary data scarcity. Experiments show superior performance, particularly in few-shot scenarios, with efficient computation. This work advances multimodal fusion for time series, offering a practical solution for real-world forecasting.
\bibliography{aaai2026}
\end{document}